\documentclass[runningheads,a4paper]{llncs}

\usepackage{amssymb}
\usepackage{graphicx}

\usepackage{url}
\urldef{\mailsa}\path{ mathcci@yahoo.com}
\urldef{\mailsb}\path {dvh@vnu.edu.vn}
\urldef{\mailsc}\path|erika.siebert-cole, peter.strasser, lncs}@springer.com|

\begin{document}

\mainmatter  % start of an individual contribution

% first the title is needed
\title{Robust Face Recognition using Local Illumination Normalization and Discriminant Feature Point Selection}%

% a short form should be given in case it is too long for the running head
\titlerunning{Robust Face Recognition using Local Illumination Normalization ...}

% the name(s) of the author(s) follow(s) next
%
% NB: Chinese authors should write their first names(s) in front of
% their surnames. This ensures that the names appear correctly in
% the running heads and the author index.
%
\author{Song Han, Jinsong Kim, Cholhun Kim, Jongchol Jo, and Sunam Han }
\authorrunning{Song Han, Jinsong Kim, Cholhun Kim, Jongchol Jo, and Sunam Han}
% (feature abused for this document to repeat the title also on left hand pages)

% the affiliations are given next; don't give your e-mail address
% unless you accept that it will be published
\institute{Faculty of Mathematics, Kim Il Sung University, D.P.R.K\\
\mailsa
}
%\institute {$^3}$ Faculty of Information Technology, Vietnam National University, Vietnam\\
%\mailsb}

%
% NB: a more complex sample for affiliations and the mapping to the
% corresponding authors can be found in the file "llncs.dem"
% (search for the string "\mainmatter" where a contribution starts).
% "llncs.dem" accompanies the document class "llncs.cls".
%

\toctitle{Lecture Notes in Computer Science}
\tocauthor{Authors' Instructions}
\maketitle

\begin{abstract}
Face recognition systems must be robust to the variation of various factors such as facial expression, illumination, head pose and aging. Especially, the robustness against illumination variation is one of the most important problems to be solved for the practical use of face recognition systems. Gabor wavelet is widely used in face detection and recognition because it gives the possibility to simulate the function of human visual system. In this paper, we propose a method for extracting Gabor wavelet features which is stable under the variation of local illumination and show experiment results demonstrating its effectiveness.   
\end{abstract}

\section{Introduction}
Image-based face recognition systems can be categorized into two categories, that is, holistic based recognition system and feature-based recognition system [1-2]. In this section, we consider previous works which were proposed to suppress the influence of illumination variation in feature-based face recognition. 

Face recognition based on the local information such as Gabor wavelet features in feature points has advantage that recognition is not sensitive to the variation of head pose or illumination, compared to the holistic based recognition such as Eigenfaces, Fisherfaces, ICA(Independent Component Analysis) and HMM(Hidden Markov Model) [4-7]. Different models and methods were proposed to support local feature-based face recognition, such as AAM(Active Appearance Model), LBP(Local Binary Pattern) and EBGM(Elastic Bunch Graph Matching) [8-18]. AAM is a geometrical and statistical model of face image, which is obtained by applying PCA or LDA to feature vectors consisting of coordinate and gray level intensity of feature points. LBP is a method to represent face image using binary vectors corresponding to each pixel, which are constructed by comparing pixel value of each pixel enclosing a pixel with the value of center pixel. Many training data and time are needed to construct AAM or LBP-based classifier. 

EBGM is a matching method using Gabor wavelet features extracted from the face graph reflecting topological structure of human faces, where face graph is constructed by connecting landmarks (node) of face with edges [14]. Each node of face graph has 40-dimensional Gabor wavelet coefficient vector corresponding to 5 frequencies and 8 directions.  EBGM-based face recognition has drawback that estimation of landmarks and extraction of Gabor wavelet coefficients are influenced by the illumination variation. To overcome such a weakness of EBGM, some researchers tried to develop techniques which can decrease the effect of illumination variation by applying normalization or DCT(Discrete Cosine Transform) to the input image [16-18]. However, it is still difficult to remove the effect of local illumination variation simultaneously with global illumination, though their methods make it possible to suppress the effect of global illumination variation to a certain degree.

In this paper, we introduce a method extracting Gabor wavelet features and discriminant feature points from face image, which is robust against local illumination variation and shows high recognition efficiency. FERET and YaleB are used to demonstrate the effectiveness of our methods.

\section{Extraction of Robust Gabor Wavelet Features}

In this section, we describe a method extracting Gabor wavelet features from face images, which is robust against local illumination variation. Let $I(\overrightarrow{x})$  be an face image, where $\overrightarrow{x}=(x,y)\in X \times Y$. Gabor wavelet transformation of $I(\overrightarrow{x})$  is defined as follows [14, 16]. 

\begin{equation}
G_j(\overrightarrow{x})=\int_{X \times Y}I(\overrightarrow{x}^{'})\Psi_j(\overrightarrow{x}-\overrightarrow{x}^{'}) d^{2}\overrightarrow{x}^{'}
\end{equation}					                                                          

Here,  
\[\begin{array}{llll}
\Psi_j(\overrightarrow{x})=\frac{\overrightarrow{k}_j^2}{\sigma^2}\exp(-\frac{\overrightarrow{k}_j^{2}\overrightarrow{x}^2}{2\sigma^2})[\exp(i\overrightarrow{k}_j\overrightarrow{x})-\exp(-\frac{\sigma^2}{2})],\\
\end{array}\]

\[\begin{array}{llll}
\overrightarrow{k}_j = \left(
\begin{array}{ccc}
k_{jx}   \\
k_{jy}   
\end{array} \right)
=\left(
\begin{array}{ccc}
k_{v}cos\phi_{\mu}   \\
k_{v}sin\phi_{\mu} 
\end{array} \right),
\end{array}\]

\[\begin{array}{llll}
k_v=2^{-(v+2)/2} (v=0,...,4), \phi_{\mu} = \mu \frac{\pi}{8} (\mu=0, ..., 7), 
\end{array}\]

\[\begin{array}{llll}
j=\mu + 8v (0 \leq j \leq 39 = m).
\end{array}\]
The second term in the integral of $\Psi_j(\overrightarrow{x})$  makes the kernel DC-free, that is, $ \int \Psi_j (\overrightarrow{x}) d^{2}\overrightarrow{x}=0$. The vector consisting of coefficients of Gabor wavelet transformation corresponding to each orientation and frequency for each point of face image is called jet and denoted by  $ g(\overrightarrow{x})=(G_1(\overrightarrow{x})), ..., G_m(\overrightarrow{x}))=(G_j(\overrightarrow{x}))_{j=0}^m$.   $G_j(\overrightarrow{x})$ can be calculated as shown in Eq.2 by considering the fact that   $G_j(\overrightarrow{x})$ only depends on the points belonging to the neighborhood   $D_j(\overrightarrow{x})$ of $\overrightarrow{x}$  from the property of kernel function $\Psi_j$  and the characteristics of the calculation of Gabor wavelet coefficients. 

\begin{equation}
G_j(\overrightarrow{x})=\int_{D_j(\overrightarrow{x})}I(\overrightarrow{x}^{'})\Psi_j(\overrightarrow{x}-\overrightarrow{x}^{'}) d^{2}\overrightarrow{x}^{'}
\end{equation}	 						                                        

$D_j(\overrightarrow{x})$  is a small area compared to the input face image and therefore we can assume that gray level intensity of face image changes linearly according to the brightness and contrast of illumination in this area, i.e.,

\begin{equation}
I(\overrightarrow{x}^{'})=\alpha_{bj} (\overrightarrow{x})+\alpha_{cj}(\overrightarrow{x})\cdot I_{D_j(\overrightarrow{x})}^{'} (\overrightarrow{x}^{'}), (\overrightarrow{x}^{'}\in D_j(\overrightarrow{x}))
\end{equation}

Here,  $\alpha_{bj} (\overrightarrow{x})$ and $\alpha_{cj}(\overrightarrow{x})$ are the average and standard deviation of gray level intensity in local area  $D_j (\overrightarrow{x})$, and express brightness and contrast of illumination. From Eq.2 and Eq.3, we can calculate   $G_j (\overrightarrow{x})$ as follows. \\
\\

$G_j (\overrightarrow{x})=\int_{D_i(\overrightarrow{x})}I(\overrightarrow{x}^{'})\Psi_j(\overrightarrow{x}-\overrightarrow{x}^{'}) d^{2}\overrightarrow{x}^{'}=$\\

$\int_{D_j(\overrightarrow{x})}  (\alpha_{bj} (\overrightarrow{x})+\alpha_{cj}(\overrightarrow{x})\cdot I_{D_j(\overrightarrow{x})}^{'} (\overrightarrow{x}^{'}))\cdot    \Psi_j(\overrightarrow{x}-\overrightarrow{x}^{'}) d^{2}\overrightarrow{x}^{'}=$\\

$\alpha_{bj} (\overrightarrow{x}) \int_{D_j(\overrightarrow{x})}  \Psi_j(\overrightarrow{x}-\overrightarrow{x}^{'}) d^{2}\overrightarrow{x}^{'}$+\\

$\alpha_{cj} (\overrightarrow{x})\int_{D_j(\overrightarrow{x})} I_{D_j(\overrightarrow{x})}^{'} (\overrightarrow{x}^{'})) \Psi_j(\overrightarrow{x}-\overrightarrow{x}^{'}) d^{2}\overrightarrow{x}^{'}=\alpha_{bj} (\overrightarrow{x}) \Phi_j + \alpha_{cj} (\overrightarrow{x}) G_{j0}(\overrightarrow{x})$\\
\\

Here,  $G_{j0}(\overrightarrow{x})=\int_{D_j(\overrightarrow{x})} I_{D_j(\overrightarrow{x})}^{'} (\overrightarrow{x}^{'})) 
\Psi_j(\overrightarrow{x}-\overrightarrow{x}^{'}) d^{2}\overrightarrow{x}^{'}$  and  $\Phi_j=\int_{D_j(\overrightarrow{x})} \Psi_j(\overrightarrow{x}$\\
$-\overrightarrow{x}^{'}) d^{2}\overrightarrow{x}^{'}$, because the size of $D_j(\overrightarrow{x})$ changes depending on only  $j$, not $\overrightarrow{x}$. Thus, the following equation holds.

\begin{equation}
G_j (\overrightarrow{x})=\alpha_{bj} (\overrightarrow{x}) \Phi_j + \alpha_{cj} (\overrightarrow{x}) G_{j0}(\overrightarrow{x})
\end{equation}

We can think that $G_{j0}(\overrightarrow{x})$  is not influenced by the illumination variation because local illumination variation will appears only as the change of values of $\alpha_{bj} (\overrightarrow{x})$ and $\alpha_{cj}(\overrightarrow{x})$. By using  $G_{j0}(\overrightarrow{x})$ as the Gabor wavelet coefficients instead of  $G_j (\overrightarrow{x})$, we can achieve the effect of normalization of local illumination variation.   $G_{j0}(\overrightarrow{x})$ can be calculated as follows. 

\begin{equation}
G_{j0} (\overrightarrow{x})=(G_j (\overrightarrow{x})-\alpha_{bj} (\overrightarrow{x})\Phi_j)/\alpha_{cj} (\overrightarrow{x})
\end{equation}
  			                                                                          
\section{Feature Point Selection using Separability of Classes}

In [19], authors proposed a method for selecting feature points using discriminant vector, but it is hard to say that it gives most discriminant points for Gabor wavelet features. Linear discriminant analysis is a method to find out subspace in which the separability of classes becomes maximum, where the separability of classes is defined as the ratio of between-class variation to within-class variation [20-23]. It can be considered that the points having big separability of classes are the most discriminant points when selecting feature point. We calculate separability of classes using Gabor wavelet coefficient vector $g(\overrightarrow{x})$  for each point and select the points whose separability of classes are big as feature points for the extraction of jet. Here, Gabor wavelet coefficients only use magnitude and neglect phase component, i.e.,  $g(\overrightarrow{x})=(|G_{j0}(\overrightarrow{x})|)_{j=0}^m$. 

Let $\hat{S}_W(\overrightarrow{x})$  and $\hat{S}_b(\overrightarrow{x})$  be the within-class/between-class scatter matrix by considering feature vector $g(\overrightarrow{x})$ of fixed point $\overrightarrow{x}$  as a random vector. Then, the value $\hat{J}(\overrightarrow{x})$ of the separability of classes for each $\overrightarrow{x}\in X \times Y$ is calculated as follows [20]. 

\begin{equation}
\hat{J}(\overrightarrow{x})=\frac{tr(\hat{S}_b(\overrightarrow{x}))}{tr(\hat{S}_W(\overrightarrow{x}))}
\end{equation}

We select $P_F=\{ \overrightarrow{x}_1,..., \overrightarrow{x}_N|\hat{J}(\overrightarrow{x}_i)>\varepsilon_0, i=1,...,N \} $ as the candidate set for feature points, where $\varepsilon_0$  is a threshold. Absolute value $|G_{j0}(\overrightarrow{x})|$ of Gabor wavelet coefficient does not change largely in the neighborhood of $\overrightarrow{x}$  and thus correlation between feature vectors becomes large and value of  $\hat{J}(\overrightarrow{x})$ becomes similar. This implies that the points of $P_F$  are mainly placed in the neighborhoods of extreme points of the function $\hat{J}(\overrightarrow{x}): X \times Y \rightarrow R$. 
To increase recognition rate, however, it is important not only to select feature points having big separability of classes but also to make the correlation between feature vectors small as possible. Thus, we finally select the points from $P_F$, which have a less correlation each other, using correlation between feature vectors and k-means clustering algorithm. Selection algorithm is describe below. 

Let  $q(1<q<N)$, $l$ and $c$  denote the number of groups, the number of step and the maximum repeat count. Let $P_k^{(l)} (k=1,..., q)$  be the  $k$-th group in step $l$. Here,  $P_k^{(l)}\subset P_F$. And let   $x_k^{(l)}$be the center of $k$-th group in stpe  $l$. \\

$\emph{Step1}$: Initialize parameters  $l$ and  $x_k^{(0)}$ as   $l=0$ and   $x_k^{(0)}=\overrightarrow{x}_k (k=1,...,q)$. 

$\emph{Step2}$: Reassign points $\overrightarrow{x}_i(i=1, ..., N)$  to group  $P_{k_i}^{(l)}$. Here,  $k_i$  is decided as follows. 
\begin{equation}
k_i=argmax_{j} s_{ij}, s_{ij}=s(\overrightarrow{x}_i, P_k^{(l)})=E\{\frac{<g(\overrightarrow{x}_i), g(x_j^{(l)})>}{||g(\overrightarrow{x}_i)|| \cdot ||g(x_j^{(l)}||}\}
\end{equation}

$\emph{Step3}$: If $l<c$  and there is a point $\overrightarrow{x}_i (1\leq i \leq N)$  which is reassigned, then increase $l$  one and recalculate $x_k^{(l)}$  and go to step 2. The center $x_k^{(l)}$  of group $P_k^{(l)}$  is calculated using the equation described below, where $P_k^{(l)}=\{\overrightarrow{x}_{i,1}, ..., \overrightarrow{x}_{i,q_k}\}$.

 \begin{equation}
x_k^{(l)}=\frac{\Sigma_{j=1}^{q_k} \hat{j}(\overrightarrow{x}_{k,j})\cdot \overrightarrow{x}_{k,j}}{\Sigma_{j=1}^{q_k} \hat{j}(\overrightarrow{x}_{k,j})}
\end{equation}        
              
\emph{Step4}: If the above condition is not satisfied, then $P_F^{'}=\{x_i^{l}, ..., x_q^{(l)}\}$  is the set of feature points. Finish algorithm.

\begin{table}
\centering
\includegraphics[height=3.8cm]{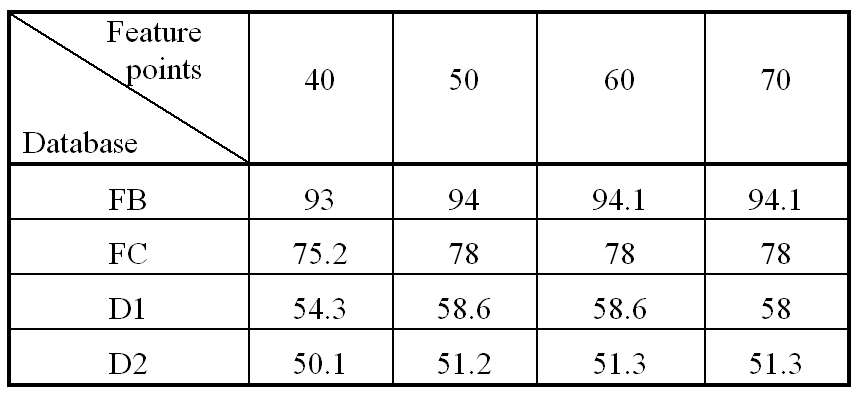}
\caption{Recognition performance according to the number of feature points ($\%$)}
\label{fig:example}
\end{table}

\section{Performance Estimation}

FA, FB, FC, D1, D2 (3539 images) of FERET and YaleB (4500 face images, 10 persons) are used for experiment. The purpose of the experiment is to confirm that recognition performance is increased in FC under illumination variation and YaleB, but don’t decrease in FB, D1 and D2, compared to the previous methods, when we apply the method of the paper. For the experiment, 128$\times$128 image extracted from the image where the position of eyes are given. The sum of similarities between jets is used as the similarity for performance estimation [14]. 

Table 1 shows the recognition performance according to the number of feature points, when the method of paper is applied. As we can see in the table, recognition performance does not increase when the number of feature points is greater than 50. 

\begin{table}
\centering
\includegraphics[height=2.8cm]{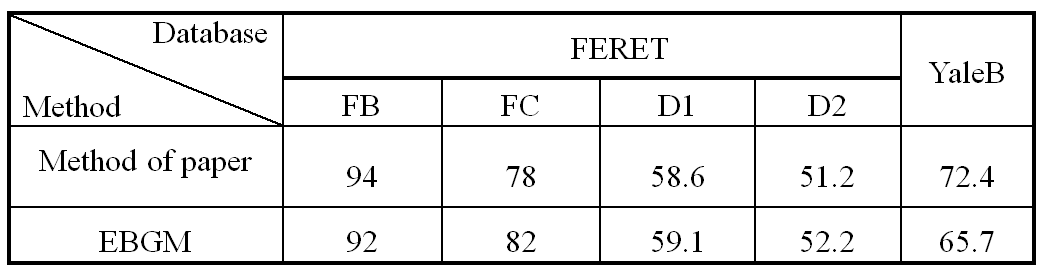}
\caption{Comparison of recognition performance with EBGM ($\%$)}
\label{fig:example}
\end{table}

Table 2 shows the result of comparison between the method of paper and EGBM in case that the number of feature points is 50. As we can know from the table, it is possible to remove performance degradation which is caused by the incorrect extraction of feature points under serious illumination variation (YaleB), if we apply the method of paper. However, if we fix landmark, the performance is decreased in case that there is no serious illumination variation.

Table 3 shows recognition performances in case that we use $G_j(\overrightarrow{x})$ and $G_{j0}(\overrightarrow{x})$ as wavelet feature coefficients respectively. As we can see in the table, normalization of the paper demonstrates high recognition performance without being affected by the illumination variation even in case that there is a change of facial expression or aging . 
\begin{table}
\centering
\includegraphics[height=3.0cm]{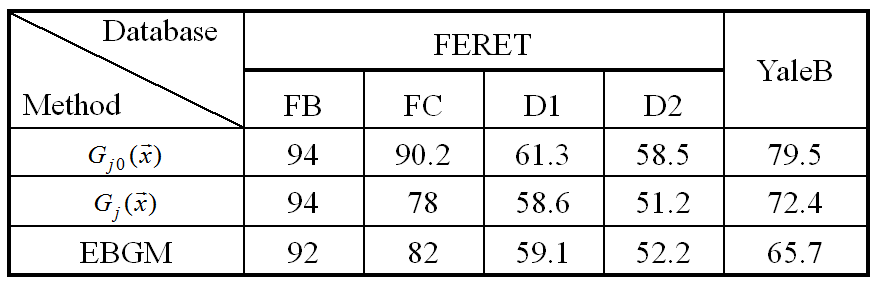}
\caption{Recognition performance under normalization($\%$)}
\label{fig:example}
\end{table}

\end{document}